\documentclass[letterpaper, 10 pt, conference]{ieeeconf}
\usepackage{times}
\usepackage{graphicx}
\usepackage{amsmath,amssymb,amsopn,amstext,amsfonts}
\usepackage{pdfsync}
\usepackage{balance}
\usepackage{color}
\usepackage{algorithm}
\usepackage{multirow}
\usepackage{algorithmic}
\usepackage{epstopdf}
\usepackage{mathtools}
\usepackage{float}
\usepackage{subfig}
\usepackage{hyperref}

\graphicspath{{figures/}}
\DeclareGraphicsExtensions{.pdf,.png,.jpg}
\IEEEoverridecommandlockouts
\title{\LARGE \bf High-Precision Online Markerless Stereo Extrinsic Calibration}

\author{Yonggen Ling and Shaojie Shen
	\thanks{All authors are with the Department of Electronic and Computer Engineering, The Hong Kong University of Science and Technology, Hong Kong, China. {\tt\small ylingaa@connect.ust.hk, eeshaojie@ust.hk}}
	\thanks{This work was supported by HKUST project R9341. } 
}

\begin{document}

\maketitle
\thispagestyle{empty}
\pagestyle{empty}

%%%%%%%%%%%%%%%%%%%%%%%%%%%%%%%%%%%%%%%%%%%%%%%%%%%%%%%%%%%%%%%%%%%%%%%%%%%%%%%%
\begin{abstract}
Stereo cameras and dense stereo matching algorithms are core components for many robotic applications due to their abilities to directly obtain dense depth measurements and their robustness against changes in lighting conditions.
However, the performance of dense depth estimation relies heavily on accurate stereo extrinsic calibration.
In this work, we present a real-time markerless approach for obtaining high-precision stereo extrinsic calibration using a novel 5-DOF (degrees-of-freedom) and nonlinear optimization on a manifold, 
which captures the observability property of vision-only stereo calibration. 
Our method minimizes epipolar errors between spatial per-frame sparse natural features.
It does not require temporal feature correspondences, making it not only invariant to dynamic scenes and illumination changes, 
but also able to run significantly faster than standard bundle adjustment-based approaches. 
We introduce a principled method to determine if the calibration converges to the required level of accuracy, and show through online experiments that our approach achieves a level of accuracy that is comparable to offline marker-based calibration methods. 
Our method refines stereo extrinsic to the accuracy that is sufficient for block matching-based dense disparity computation.
It provides a cost-effective way to improve the reliability of stereo vision systems for long-term autonomy. 
\end{abstract}

%%%%%%%%%%%%%%%%%%%%%%%%%%%%%%%%%%%%%%%%%%%%%%%%%%%%%%%%%%%%%%%%%%%%%%%%%%%%%%%%
\section{Introduction}
\label{sec:introduction}
Stereo cameras and dense stereo vision have become one of the most widely used sensing modalities for robotic applications.
High-accuracy calibration of parameters, including the intrinsic of each camera and stereo extrinsic, is the key to obtain high-quality and dense depth measurements.
A good stereo calibration enables transformations of stereo images such that the epipolar lines become parallel.
This is the foundation for most dense stereo matching algorithms, as dense stereo correspondences can be simplified to efficient 1D-search problems on epipolar lines.
Traditionally, stereo cameras are mounted rigidly, utilizing advanced materials and mechanical structures, and stereo calibration is conducted offline using known planar patterns~\cite{tsai87, zhengyouZhang00}. 
However, the calibration parameters tend to change over time due to changes in temperature, vibration, or unexpected shocks to the system.
The ability for stereo cameras to perform online self-calibration thus becomes a key factor to enable long-term autonomy for robots using a stereo-based sensing modality.

In recent years, markerless stereo self-calibration has been of growing interest to the research community. 
Self-calibration algorithms automatically determine the intrinsic and extrinsic parameters of stereo cameras that are equipped on a fixed stereo baseline, without the necessity of manual recalibration. 
By utilizing self-calibration, industrial products based on stereo cameras can be more practical in the long run and more user-friendly since they do not need tedious re-calibration of stereo cameras using a known target.

\begin{figure}[t]
	\centering
	\subfloat[]{{\includegraphics[width=4.1cm]{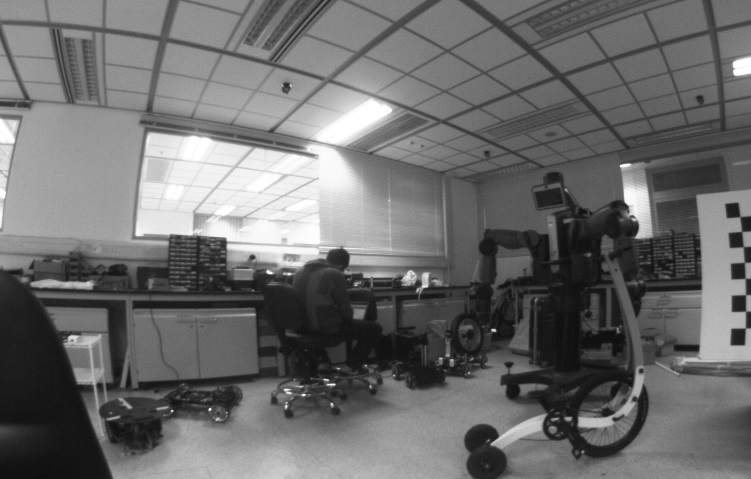} }}
	\subfloat[]{{\includegraphics[width=4.1cm]{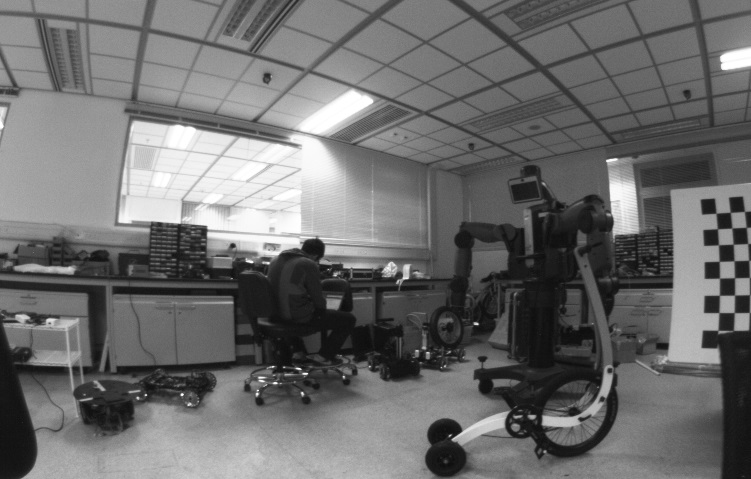} }} \\
	\subfloat[]{{\includegraphics[width=4.1cm]{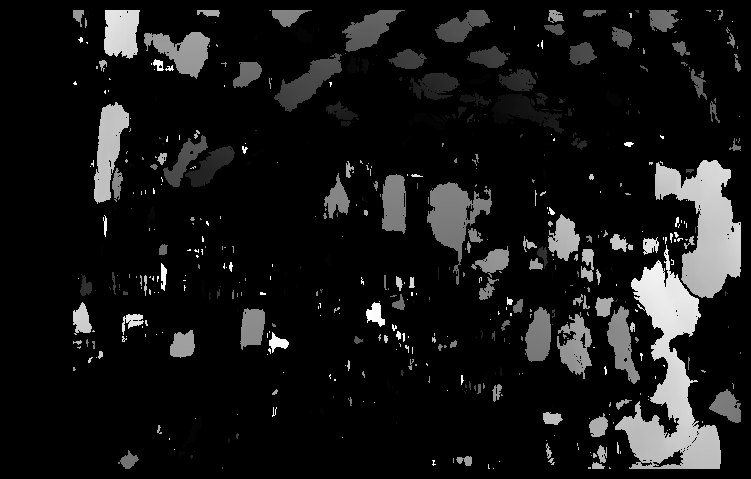} }}
	\subfloat[]{{\includegraphics[width=4.1cm]{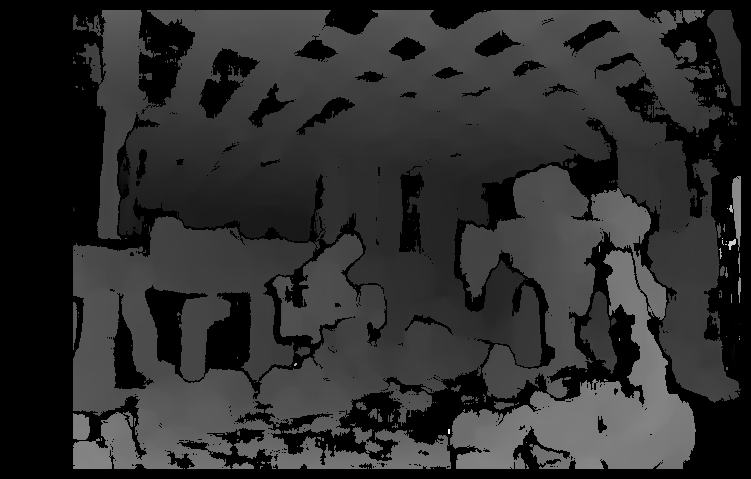} }}
	\caption{(a) and (b) are raw left and right images. 
	         (c) is the disparity map computed using the initial stereo extrinsic with errors up to 3 degrees in each of the rotating axes. 
	         (d) is the disparity map computed using extrinsic parameters refined by our proposed algorithm. 
	         The disparity map is obtained via a standard block matching algorithm (BM in OpenCV) after stereo rectification (stereoRectify in OpenCV).}
	\label{fig:calibration}
\end{figure}

Based on studies performed using different types of stereo cameras, we identify that the cameras' intrinsic parameters do not change noticeably even in the event of crashing or long-term mechanical vibration.
However, the stereo extrinsic, which is the relative transformation between two cameras, changes significantly even with moderate shocks or after lengthy operations on a vibrating system such as aerial robots.
Wide baseline stereo cameras suffer even more extrinsic deviations. 

As shown in Fig.~\ref{fig:calibration}, even minor errors in the stereo extrinsic can result in significant performance loss in dense stereo matching due to misalignment between the epipolar lines.
We therefore focus on addressing the problem of real-time stereo extrinsic calibration without using artificial markers. 
We assume that an initial guess of extrinsic parameters is available with orientation errors up to several degrees. 
We also assume that the length of the stereo rig is constant. This makes sense since the material (usually metal) used to build the stereo mount cannot be compressed.
Even if this assumption is invalid, the depth error increases linearly with respect to the error in the baseline length, which is very minor for long-range depth preception.
Furthermore, the property that incorrect baseline length does not add any invalid pixels to the disparity map is very desirable, as the percentage of valid pixels in the disparity map translates directly into the obstacle detection capability, which determines the performance of fundamental autonomy modules such as obstacle avoidance and motion planning.
In our formulation, this constant baseline length assumption helps to recover the scale factor in our vision-only stereo calibration case.

Our goal is to perform online refinement of stereo extrinsic parameters. 
We aim to achieve a precision level such that the calibrated stereo images can be directly input to standard dense stereo matching algorithms based on 1D epipolar line search.
To this end, we identify our contributions as follows:
\begin{itemize}
	\item We propose a novel 5-DOF nonlinear optimization on a manifold for estimating the stereo extrinsic parameteres using natural features in both static and dynamic scenes.
	\item We derive a mathematical expression for computing the covariance of stereo extrinsic estimates to identify when the calibration result is sufficiently accurate.
	\item We present careful engineering decisions on feature selection, outlier rejection, and the system pipeline to achieve high-precision stereo extrinsic calibration that is suitable for dense block matching-based disparity computation.
	\item Our method runs real-time onboard moderate computers. Extensive online experiments are presented to show that the calibration performance is comparable to standard marker-based offline methods.
\end{itemize}

The rest of the paper is structured as follows. In Sect.~\ref{sec:literature}, we review the state-of-the-art scholarly work. 
The methodology and system are discussed in Sect.~\ref{sec:methodology}. 
Implementation details and experimental evaluations are presented in Sect.~\ref{sec:experiment}. 
Sect.~\ref{sec:conclusions} draws the conclusions and points out possible future extensions.

\section{Related Work}
\label{sec:literature}
There is an extensive collection of scholarly work related to stereo calibration. 
The majority of stereo calibration algorithms determine the cameras' intrinsic and extrinsic by observing known targets with overlapping views \cite{tsai87, zhengyouZhang00}. 
Bundle adjustment is then applied to refine the cameras' parameters as well as the 3D points' positions to minimize reprojection errors. 
These algorithms typically run offline due to heavy computational demands.

Stereo self-calibration without any known patterns or markers has attracted interests in recent years \cite{michael13, peter12, dang09, michael13rss, mayi01}, 
and is still an active research area because of the massive demands from different vision-based applications. 
Multiple geometric constraints, including the epipolar constraint, trilinear constraint, and bundle adjustment, 
are discussed and combined into a unified framework in \cite{dang09}. 
The calibration results are then incorporated into an iterated extended Kalman filter (EKF). 
The main drawback of \cite{dang09} is the fast growing computational complexity when the number of correspondences increases. 
To accelerate the speed of bundle adjustment optimization, \cite{michael13} presents a form of variable partition that reduces computation. 
\cite{michael13rss} relaxes the stereo transform during bundle adjustment by introducing a log barrier cost function. 
However, we stress that the observability property of \cite{michael13rss} is controversial since vision-only algorithms are not able to capture the real-world scale if the baseline length is not fixed. 
On the other hand, \cite{peter12, fivepoint, marten02} only consider the epipolar constraints to obtain the 5-DOF extrinsic. 
An approximated essential matrix is firstly computed by linear optimization, and is then refined by a nonlinear step incorporating depth constraints \cite{marten02} or is used to extract the relative extrinsic parameters \cite{fivepoint}. 
Real-time operations at the video-frame rate are reported in \cite{peter12}.

Other approaches to perform relative extrinsic calibration are achieved by means of visual-inertial fusion \cite{HesKotBow1402, LiMou1305, dong12, heng14, yangShen15}. 
Since the measurements at different locations exhibit different behaviors, relative poses of the cameras can be estimated via minimizing the visual-inertial fusion error metric. 
However, the main focus of visual-inertial fusion is not on stereo calibration. 
The relative poses of cameras are merely the by-products. 
The relative transformation between cameras is not directly modeled, but rather calculated through the concentration of two relative poses between one camera and one IMU. 
Therefore, the calibration accuracy obtained by visual-inertial methods is not accurate enough for dense disparity computation.

We decided to use epipolar error as the only error metric for the reason that epipolar constraints are generalizable to dynamic scenes.
There is no need to maintain temporal relations between feature points. 
Besides this, as mentioned in \cite{dang09}, the epipolar constraint is the most significant metric. 
Removing temporal data association brings us the benefit of being able to process a large number of feature observations without sacrificing the real-time requirement.
Most importantly, our error metric leads to calibration accuracy that is comparable to that of offline calibration. 
Most similar to our proposed approach are the methods proposed in \cite{peter12, mayi01}, which parameterize the 5-DOF variables using Euler angles or in the Stiefel space. 
We, however, parameterize them on the manifold that is tailored for stereo extrinsic estimation, which leads to simpler implementation, faster convergence, and higher accuracy. 
Moreover, the covariance of the extrinsic estimates reports the up-to-date estimation accuracy, which helps to determine the calibration termination criteria.
We also present a carefully designed and integrated real-time system with extensive performance verification. 
%\todo{Done up to this point}

\section{Methodology}
\label{sec:methodology}
\subsection{Prerequisites}
\begin{figure}[h]
	\centering
	\includegraphics[width=0.6\columnwidth]{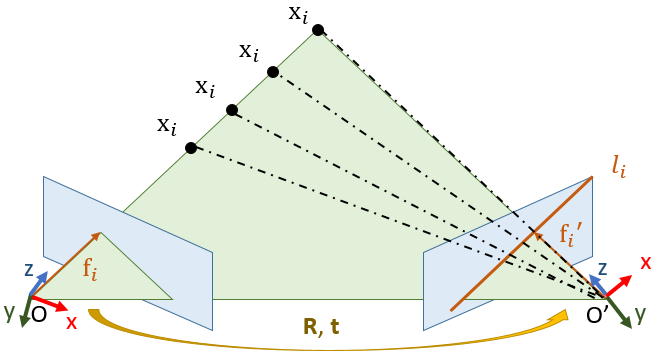}
	\caption{Illustration of a stereo epipolar system.} 
	\label{fig:notation}
\end{figure}
We begin by defining notations (also see Fig.~\ref{fig:notation}). We denote the camera matrices of the left and right cameras as $\mathbf{K} \in \mathbb{R}^{3 \times 3} $ and $\mathbf{K}' \in \mathbb{R}^{3 \times 3} $ respectively. We assume that $\mathbf{K}$ and $\mathbf{K}'$ are known and estimated by monocular camera calibration beforehand. Suppose a feature $i$ is observed both in the left image and right image. Its coordinates in the coordinate system of the left and right camera optical centers are $\mathbf{x}_i \in \mathbb{R}^{3} $ and $\mathbf{x}_i' \in \mathbb{R}^{3}$ respectively, and its projections on the left and right image are ($u_i$, $v_i$) and ($u_i'$, $v_i'$). The corresponding depths are $\lambda_i$ and $\lambda_i'$ respectively. Assuming a pinhole model of the cameras, we denote the normalized pixel coordinates of $\mathbf{x}_i$ and $\mathbf{x}_i'$ as

\begin{flalign}
\mathbf{f}_i =  \mathbf{K}^{-1} \begin{bmatrix}
u_i \\ 
v_i \\ 
1
\end{bmatrix}, \ \ \ \mathbf{f}_i' =  {\mathbf{K}'}^{-1} \begin{bmatrix}
u_i' \\ 
v_i' \\ 
1
\end{bmatrix},
\end{flalign}
where $\mathbf{x}_i = \lambda_i \mathbf{f}_i$, $\mathbf{x}_i' = \lambda_i' \mathbf{f}_i'$.
Since the coordinate systems of the left and right cameras are related by a rotation $ \mathbf{R} \in \mathbb{R}^{3 \times 3} $ and a translation $\mathbf{t} \in \mathbb{R}^{3} $, we have 
\begin{flalign}
\mathbf{x}_i' = \mathbf{R} \mathbf{x}_i + \mathbf{t}.
\end{flalign}
The essential matrix $\mathbf{E} $ is defined as \cite{richard2nd}
\begin{flalign}
\label{equ:essential}
\mathbf{E} =  \lfloor \mathbf{t} \times \rfloor \mathbf{R},
\end{flalign}
where $\lfloor \cdot \times \rfloor$ is a skew-symmetric operator that transforms a vector $\mathbf{a} = [a_1 \ a_2 \ a_3 ]^T\in \mathbb{R}^{3}$ into a skew-symmetric matrix.

For a point along the $\mathbf{f}_i$ direction in the three dimensional space of the left camera's coordinate system, its projection lies on a line $l_i$ of the right-image plane. This line $l_i$ is called the epipolar line. Mathematically, the epipolar line $l_i$ can be written as the product of essential matrix $\mathbf{E}$ and $\mathbf{f}_i$,
\begin{flalign}
l_i = \mathbf{E} \mathbf{f}_i,
\end{flalign} 
and the epipolar constraint is
\begin{flalign}
\label{equ:epi}
(\mathbf{f}_i')^Tl_i = (\mathbf{f}_i')^T \mathbf{E} \mathbf{f}_i = 0.
\end{flalign} 
Detailed treatment of epipolar geometry can be found in \cite{richard2nd}.

\subsection{Formulation}
\label{sec:parameterization}
Since the locations of detected feature points are corrupted by noise, whose source is pixel coordinate quantization and limited detector performance, they do not satisfy the epipolar constraint perfectly. The epipolar error for each pair of matched feature points $\mathbf{f}_i$ and $\mathbf{f}_i'$ is defined as
\begin{flalign}
 || ( \mathbf{f}_i')^T \mathbf{E} \mathbf{f}_i ||^2,
\end{flalign} 
which is the square of the distance between point $\mathbf{f}_i'$ and epipolar line $l_i$ in view of epipolar geometry. Our goal is to solve for $\mathbf{R}$ and $\mathbf{t}$ so that the overall epipolar error is minimized for all the matched feature correspondences:
\begin{flalign}
\label{equ:cost}
&\underset{ \mathcal{ \mathbf{R}, \mathbf{t}} }{\min}  \ \ \sum_{i=0}^{N} || ( \mathbf{f}_i')^T  \lfloor \mathbf{t} \times \rfloor \mathbf{R} \mathbf{f}_i ||^2, \\
& \mbox{s.t.} \ \ \ \ ||\mathbf{t}|| = 1,
\end{flalign}
where $\mathbf{E} $ is substituted by $\lfloor \mathbf{t} \times \rfloor \mathbf{R} $ using equation~\eqref{equ:essential}.
\begin{figure}[H]
	\centering
	\includegraphics[width=0.6\columnwidth]{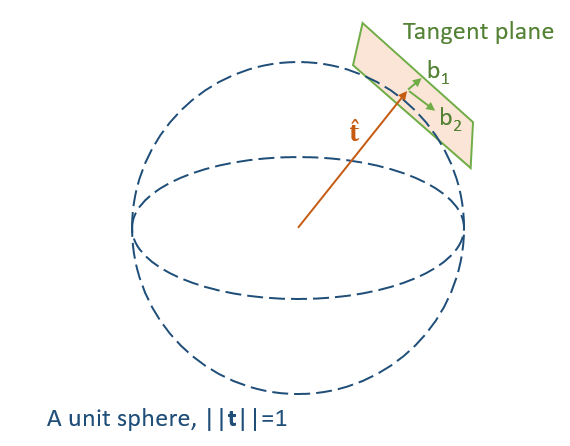}
	\caption{Since the translation $\mathbf{t}$ is an up-to-scale translation in the 3D space, the degrees-of-freedom of $\mathbf{t}$ is 2. As shown in the figure, $\mathbf{t}$ lies on a unit sphere. For small disturbance on the tangent plane of current estimate $\hat{\mathbf{t}}$,  $\mathbf{t} = \hat{\mathbf{t}} + \alpha \mathbf{b}_1 + \beta \mathbf{b}_2$, where $\mathbf{b}_1$ and $\mathbf{b}_2$ are two orthogonal bases on the tangent plane, and $\alpha$ and $\beta$ are the corresponding movements towards $\mathbf{b}_1$ and $\mathbf{b}_2$. } 
	\label{translation}
\end{figure}
The unit norm constraint $ ||\mathbf{t}|| = 1$ reflects the fact that the essential matrix $\mathbf{E}$ has 5 degrees-of-freedom, 3 for rotation $\mathbf{R} \in SO(3) $ and 2 for up-to-scale translation $\mathbf{t} \in \mathbb{S}^{2}$ (equation~\eqref{equ:epi} is satisfied by whatever scaling factor $\mathbf{t}$ multiplies). We apply the theory of the Lie group and Lie algebra \cite{ma2012invitation}, and parameterize the small disturbance $\delta \boldsymbol{\theta} \in \mathbb{R}^{3}$ on the tangent space of the current estimated $\hat{\mathbf{R}}$ as
\begin{flalign}
\label{equ:parameterization_R}
\mathbf{R} = \hat{\mathbf{R}} ( \mathbf{I}_3 + \lfloor  \delta\boldsymbol{\theta} \times \rfloor ) ,
\end{flalign}
where $\mathbf{I}_3$ is a 3-by-3 identity matrix. 

For small disturbance $\delta \mathbf{t} = [\alpha\,,\,\beta]^T$ on the tangent space of current estimated translation $\hat{\mathbf{t}}$, we propose to parameterize it as 
\begin{flalign}
\label{equ:parameterization_t}
\mathbf{t} = \hat{\mathbf{t}} + [\mathbf{b}_1 \ \ \mathbf{b}_2]\delta \mathbf{t} =  \hat{\mathbf{t}} + \alpha\mathbf{b}_1 +  \beta \mathbf{b}_2,
\end{flalign}
where $\mathbf{b}_1$ and $\mathbf{b}_2$ are two orthogonal bases spanning the tangent plane, as shown in Fig.~\ref{translation}, and $\alpha$ and $\beta$ are the small displacements towards bases $\mathbf{b}_1$ and $\mathbf{b}_2$ respectively.

We also propose a method to find the bases $\mathbf{b}_1$ and $\mathbf{b}_2$. As $\mathbf{b}_1$ and $\mathbf{b}_2$ lies on the tangent plane of current estimate $\hat{\mathbf{t}}$, they are perpendicular to $\hat{\mathbf{t}}$. Moreover, $\mathbf{b}_1$ and $\mathbf{b}_2$ are not unique because there are many pairs of them that span the tangent plane. One possible method is to seek for $\mathbf{b}_1$ and $\mathbf{b}_2$ such that $\mathbf{b}_1$, $\mathbf{b}_2$ and  $\hat{\mathbf{t}}$ are orthogonal to each other. This can be done by arbitrarily selecting two bases $\mathbf{b}_1$, $\mathbf{b}_2$ and then remove the dependent components one by one using Gram-Schmidt process until the orthogonal constraint is satisfied.

The proposed algorithm is summarized in Algorithm~\ref{alg:summary}. Lines 1 to 7 check which dimension of $\hat{\mathbf{t}}$ is significant. Lines 9 to 15 select two bases according to the significant dimension of $\hat{\mathbf{t}}$. Line 17 to Line 20 is the Gram-Schmidt process that removes dependency among bases. The projection function needed in Algorithm~\ref{alg:summary} is detailed in Algorithm~\ref{alg:projection}.

\begin{algorithm}[H]
	\caption{Projection( $\mathbf{u}$, $\mathbf{v}$ )}
	\begin{algorithmic}
		\RETURN $\mathbf{u} \ \cdot $ InnerProduct( $\mathbf{u}$, $\mathbf{v}$ ) / InnerProduct( $\mathbf{u}$, $\mathbf{u}$ )
	\end{algorithmic}
	\label{alg:projection}
\end{algorithm}
\begin{algorithm}[H]
	\caption{FindingBases( $\hat{\mathbf{t}}$ )}
	\begin{algorithmic}[1]
		\STATE $idx=0$  ;
		\IF{$ \mbox{Absolute}( \hat{\mathbf{t}}(1) ) < \mbox{Absolute}( \hat{\mathbf{t}}(idx) )$}
		\STATE $idx = 1$ ;
		\ENDIF
		\IF{$ \mbox{Absolute}( \hat{\mathbf{t}}(2) ) < \mbox{Absolute}( \hat{\mathbf{t}}(idx) )$}
		\STATE $idx = 2$ ;
		\ENDIF
		\STATE \;
		\IF{$idx=0$} 
		\STATE $\mathbf{b}_1 = \{0, 1, 0 \}$, $\mathbf{b}_2 = \{0, 0, 1 \}$ ;
		\ELSIF{$idx=1$}
		\STATE $\mathbf{b}_1 = \{1, 0, 0 \}$, $\mathbf{b}_2 = \{0, 0, 1 \}$ ;
		\ELSE
		\STATE $\mathbf{b}_1 = \{1, 0, 0 \}$, $\mathbf{b}_2 = \{0, 1, 0 \}$ ;
		\ENDIF
		\STATE \;
		\STATE $\mathbf{b}_1 = \mathbf{b}_1 $ - Projection(  $\hat{\mathbf{t}}$ , $\mathbf{b}_1$ ) ;
		\STATE Normalize( $\mathbf{b}_1$ ) ;
		\STATE $\mathbf{b}_2 = \mathbf{b}_2$ - Projection(  $\hat{\mathbf{t}}$ , $\mathbf{b}_2$ ) - Projection(  $\mathbf{b}_1$ , $\mathbf{b}_2$ ) ;
		\STATE Normalize( $\mathbf{b}_2$ ) ;	
		\RETURN $\mathbf{b}_1$, $\mathbf{b}_2$
	\end{algorithmic}
	\label{alg:summary}
\end{algorithm}

\subsection{Nonlinear Optimization}
\label{sec:opt}
Realizing the fact that a calibration prior is usually available from the knowledge of the hardware setup or 5-point linear method \cite{fivepoint}, we proposed to use an iterative nonlinear optimization parameterized on the $SO(3) \times \mathbb{S}^{2}$ manifold to solve  $\mathbf{E}$,  $\mathbf{R}$ and  $\mathbf{t}$. We optimize equation~\eqref{equ:epi} in one step using the parameterization proposed in Sect~\ref{sec:parameterization}. For better presentation, we define $r_i ( \mathbf{t}, \mathbf{R} ) = ( \mathbf{f}_i')^T \lfloor \mathbf{t} \times \rfloor \mathbf{R}  \mathbf{f}_i$. The first order Taylor expension of $r_i ( \mathbf{t}, \mathbf{R} )$ around the current estimate $\hat{\mathbf{t}}$, $\hat{\mathbf{R}}$ is
%Five-point method decomposes the computation into two steps. The first step is to calculate a linear approximation of essential matrix $\mathbf{E}$ by relaxing all the internal constraints that origin from equation~\eqref{equ:essential}. The second step is to determine $\mathbf{R}$ and $\mathbf{t}$ from the linear approximation of $\mathbf{E}$ obtained in the first step. As each step of this approach incurs errors, the precision of $\mathbf{R}$ and $\mathbf{t}$ estimated by five-point method is far from enough for stereo calibration. Therefore, unlike five-point method, we optimize equation~\eqref{equ:epi} in one step using the parameterization proposed in Sect~\ref{sec:parameterization}. For better presentation, we define $r_i ( \mathbf{t}, \mathbf{R} ) = ( \mathbf{f}_i')^T \lfloor \mathbf{t} \times \rfloor \mathbf{R}  \mathbf{f}_i$. The first order Taylor expension of $r_i ( \mathbf{t}, \mathbf{R} )$ around current estimate $\hat{\mathbf{t}}$, $\hat{\mathbf{R}}$ is
\begin{flalign}
\label{equ:linear_approximation}
r_i ( \mathbf{t}, \mathbf{R} ) = r_i ( \hat{\mathbf{t}}, \hat{\mathbf{R}} ) + \mathbf{J}_i( \hat{\mathbf{t}}, \hat{\mathbf{R}} ) \cdot \Delta,
\end{flalign}
where $\Delta = \{\delta\boldsymbol{\theta}, \delta \mathbf{t} \} \in \mathbb{R}^{5}$ is the error state vector and $\mathbf{J}_i( \hat{\mathbf{t}}, \hat{\mathbf{R}} ) $ is the Jacobian of $r_i ( \mathbf{t}, \mathbf{R} )$ with respect to $\Delta$ at current estimate $ \hat{\mathbf{t}}$ and $\hat{\mathbf{R}}$,
\begin{flalign}
\mathbf{J}_i( \hat{\mathbf{t}}, \hat{\mathbf{R}} ) &= [\frac{\partial \mathbf{r}_i( \hat{\mathbf{t}}, \hat{\mathbf{R}} )}{\partial \delta\boldsymbol{\theta}} \ \frac{\partial \mathbf{r}_i( \hat{\mathbf{t}}, \hat{\mathbf{R}} )}{\partial \delta \mathbf{t} } ],\\
\frac{\partial \mathbf{r}_i( \hat{\mathbf{t}}, \hat{\mathbf{R}} )}{\partial \delta\boldsymbol{\theta}} &=  -  ( \mathbf{f}_i')^T \lfloor \hat{\mathbf{t}} \times \rfloor \hat{\mathbf{R}}  \lfloor \mathbf{f}_i  \times \rfloor,\\
\frac{\partial \mathbf{r}_i( \hat{\mathbf{t}}, \hat{\mathbf{R}} )}{\partial \delta \mathbf{t} } &= [( \mathbf{f}_i')^T \lfloor \mathbf{b}_1 \times \rfloor \hat{\mathbf{R}}  \mathbf{f}_i  \ \ \  ( \mathbf{f}_i')^T \lfloor \mathbf{b}_2 \times \rfloor \hat{\mathbf{R}}  \mathbf{f}_i].
\end{flalign}
Substituting equation \eqref{equ:linear_approximation} into \eqref{equ:cost}, we have
\begin{flalign}
\label{equ:linear_cost}
\underset{ \Delta }{\min}  \ \ \sum_{i=0}^{N-1} || r_i ( \hat{\mathbf{t}}, \hat{\mathbf{R}} ) + \mathbf{J}_i( \hat{\mathbf{t}}, \hat{\mathbf{R}} ) \cdot \Delta ||^2.
\end{flalign}
Taking the derivative with respect to $\Delta$ and setting it to zero,
%\begin{flalign}
%\label{equ:linear_derivative}
%\sum_{i=0}^{N-1} [ r_i ( \hat{\mathbf{t}}, \hat{\mathbf{R}} ) + \mathbf{J}_i( \hat{\mathbf{t}}, \hat{\mathbf{R}} ) \cdot \Delta ] = 0,
%\end{flalign}
which yields a solution to the following linear equation,
\begin{flalign}
\label{equ:linear_system}
\mathbf{J}^T \mathbf{J} \Delta = -\mathbf{J}^T \mathbf{r},
\end{flalign}
where $\mathbf{J}$ is a Jacobian matrix that is formed by stacking Jacobians $\mathbf{J}_i( \hat{\mathbf{t}}, \hat{\mathbf{R}} ) $, and $\mathbf{r}$ is the corresponding vector that is formed by stacking $ r_i ( \hat{\mathbf{t}}, \hat{\mathbf{R}} )$. To alleviate the effect of outliers arising due to imperfect feature detection and matching, we weight down large residues $r_i ( \mathbf{t}, \mathbf{R} )$ using the Huber weighting norm given by 
\begin{flalign}
w_i^h =  \begin{cases}
1, \ \ \ \  &\text{ if } \ || r_i ( \mathbf{t}, \mathbf{R} )|| \leq c_t  \\
\ \frac{c_t}{||r_i ( \mathbf{t}, \mathbf{R} )||}, \ \ \ \ &\text{ otherwise }
\end{cases},
\end{flalign}
where $c_t$ is a predefined Huber norm threshold. Statistical normalization weights $w_i^n$ from \cite{mayi01} are also used, and $w_i = w_i^n \cdot w_i^h$. Equation~\eqref{equ:linear_system} can be rewritten into a weighted formulation
\begin{flalign}
\label{equ:linear_system_weight}
\mathbf{J}^T \mathbf{W} \mathbf{J} \Delta = -\mathbf{J}^T \mathbf{W} \mathbf{r},
\end{flalign}
where $\mathbf{W}$ is a diagonal matrix that encodes the weights $w_i$. The solution for $\Delta$ is simply
\begin{flalign}
\label{equ:solution}
\Delta = -(\mathbf{J}^T \mathbf{W} \mathbf{J})^{-1} \mathbf{J}^T \mathbf{W} \mathbf{r}.
\end{flalign}
We iteratively calculate \eqref{equ:solution} and update
\begin{flalign}
\hat{ \mathbf{R}} &\leftarrow \hat{ \mathbf{R}}  \cdot \mbox{exp}(\delta\boldsymbol{\theta}), \\
\hat{ \mathbf{t}} &\leftarrow \hat{ \mathbf{t}} + \alpha \mathbf{b}_1 + \beta \mathbf{b}_2,
\end{flalign}
until convergence, where exp($\delta\boldsymbol{\theta}$) is the exponential map of $\delta\boldsymbol{\theta}$ defined in Lie algebra.

%\begin{flalign}
%\underset{ \mathcal{ \mathbf{R}, \mathbf{T}} }{\min}  \ \ &\sum_{i=0}^{N} || ( \mathbf{f}_i')^T \lfloor \mathbf{t} \times \rfloor \mathbf{R}  \mathbf{f}_i ||^2
%\end{flalign}
%Suppose 
%\begin{flalign}
%\mathbf{r}_i &=  ( \mathbf{x}_i')^T \lfloor \mathbf{t} \times \rfloor \mathbf{R}  \mathbf{x}_i \\
%&= \hat{\mathbf{r}_i} + \frac{\partial \mathbf{r}_i}{\partial \boldsymbol{\delta\theta}} \boldsymbol{\delta\theta} + \frac{\partial \mathbf{r}_i}{\partial {\alpha}} {\alpha} + \frac{\partial \mathbf{r}_i}{\partial {\beta}} {\beta}
%\end{flalign}
%where,
%\begin{flalign}
%\frac{\partial \mathbf{r}_i}{\partial \boldsymbol{\delta\theta}} &= \\
%\frac{\partial \mathbf{r}_i}{\partial {\alpha}} {\alpha} &=   \\
%\frac{\partial \mathbf{r}_i}{\partial {\beta}} {\beta} &= 
%\end{flalign}
%\begin{flalign}
%\underset{ \boldsymbol{\delta\theta}, \alpha, \beta }{\min}  \ \ &\sum_{i=0}^{N} || \hat{\mathbf{r}_i} + \frac{\partial \mathbf{r}_i}{\partial \boldsymbol{\delta\theta}} \boldsymbol{\delta\theta} + \frac{\partial \mathbf{r}_i}{\partial {\alpha}} {\alpha} + \frac{\partial \mathbf{r}_i}{\partial {\beta}} {\beta}   ||^2
%\end{flalign}
%Use Gram-Schmidt process to find the basis $\mathbf{b}_1$ and $\mathbf{b}_2$. 
\begin{figure*}[t]
	\centering
	\includegraphics[width=1.2\columnwidth]{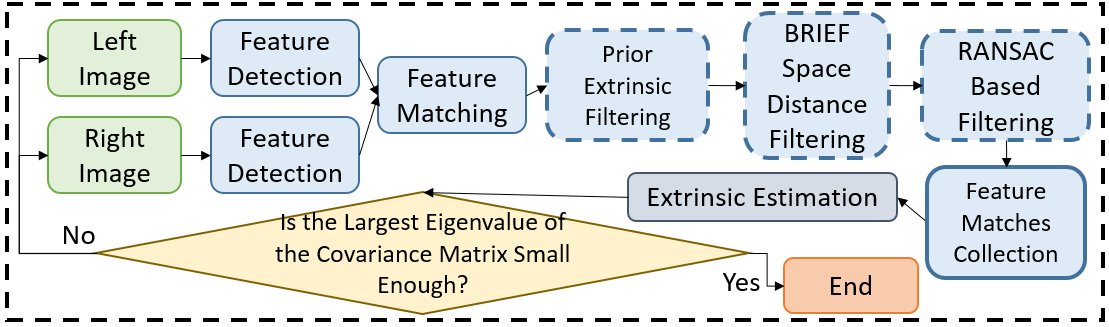}
	\caption{Overall system pipeline.} 
	\label{fig:framework}
\end{figure*}
\subsection{Estimating the Covariance for Stereo Calibration}
\label{sec:cov}
We derive the covariance of the estimates from the previous subsection mathematically. We assume that the detected features are independent and their locations in normalized pixel coordinates are corrupted by Gaussian noise with zero mean and diagonal covariance $\boldsymbol{\Sigma}_f$. That is, $\mathbf{f}_i = \mathbf{f}_i^* + \delta \mathbf{f}_i$, $\mathbf{f}_i' = \mathbf{f}_i'^* + \delta \mathbf{f}_i'$, where $\mathbf{f}_i^*$ and $\mathbf{f}_i'^*$ are the ground truth locations, and $\delta \mathbf{f}_i$ $\delta \mathbf{f}_i'$ $\sim \mathcal{N} ( \mathbf{0}, \boldsymbol{\Sigma}_f  )$. As the covariance of $\mathbf{R}$ and $\mathbf{t}$ is equal to the covariance of the error state vector $\Delta$, for the estimate $\hat{ \mathbf{t}}$ and $ \hat{ \mathbf{R}}$ in the final iteration of the optimization, we calculate the covariance for both sides of \eqref{equ:solution}, and obtain
\begin{flalign}
\Sigma_{\Delta} &= (\mathbf{J}^T \mathbf{W} \mathbf{J})^{-1} \mathbf{J}^T \mathbf{W} \Sigma_{\mathbf{r}}   ( (\mathbf{J}^T \mathbf{W} \mathbf{J})^{-1} \mathbf{J}^T \mathbf{W} )^T. \nonumber \\
&\approx \mathbf{J}^{-1} (\mathbf{J}^T \mathbf{W} )^{-1} \mathbf{J}^T \mathbf{W} \Sigma_{\mathbf{r}}  ( \mathbf{J}^{-1} (\mathbf{J}^T \mathbf{W} )^{-1} \mathbf{J}^T \mathbf{W} )^T \nonumber \\
&= \mathbf{J}^{-1} \Sigma_{\mathbf{r}}  \mathbf{J}^{-T} \nonumber \\
&= (\mathbf{J}^{T} \Sigma_{\mathbf{r}} ^{-1} \mathbf{J} )^{-1} 
\end{flalign}
where $\Sigma_{\Delta}$ is the covariance of $\Delta$, $\Sigma_{\mathbf{r}} $ is the covariance of $\mathbf{r}$, $\mathbf{J}^{-1} $ is the pseudo inverse of $\mathbf{J}$, and $(\mathbf{J}^T \mathbf{W} )^{-1}$ is the pseudo inverse of $\mathbf{J}^T \mathbf{W}$. Because the detected features are independent,  $\Sigma_{\mathbf{r}} $ is a diagonal matrix 
\begin{flalign}
\Sigma_{\mathbf{r}} &= \begin{bmatrix}
\mbox{cov}(r_0 ( \hat{\mathbf{t}}, \hat{\mathbf{R}} ))	& ... & 0 \\ 
... & \mbox{cov}(r_i ( \hat{\mathbf{t}}, \hat{\mathbf{R}} )) & ... \\ 
0 & ... & \mbox{cov}(r_{N-1} ( \hat{\mathbf{t}}, \hat{\mathbf{R}} ))
	\end{bmatrix},
\end{flalign}
where $\mbox{cov}(r_i ( \hat{\mathbf{t}}, \hat{\mathbf{R}} )) = [ ( \mathbf{f}_i')^T \hat{ \mathbf{E}} \boldsymbol{\Sigma}_f \hat{ \mathbf{E}}^T \mathbf{f}_i' + \mathbf{f}_i^T \hat{ \mathbf{E}}^T \boldsymbol{\Sigma}_f \mathbf{E}\mathbf{f}_i ]$, and $\hat{ \mathbf{E}} = \lfloor \hat{\mathbf{t}} \times \rfloor \hat{\mathbf{R}}$. The covariance of $(\delta \mathbf{f}_i')^T \hat{ \mathbf{E}} \delta \mathbf{f}_i$ is ignored since it is sufficiently small.

If fast computation is needed, we can further simplify the computation of  $\Sigma_{\Delta}$ by assuming all the $\mbox{cov}(r_i ( \hat{\mathbf{t}}, \hat{\mathbf{R}} ))$ to be the same (denoted as $c_r$) and approximate it as 
\begin{flalign}
\label{equ:approximated_cov}
\Sigma_{\mathbf{\Delta}}' &\approx  c_r (\mathbf{J}^{T} \mathbf{J} )^{-1}  \approx  c_r \cdot (\mathbf{J}^T \mathbf{W} \mathbf{J})^{-1}.
\end{flalign}
As $ \mathbf{J}^T \mathbf{W} \mathbf{J}$ is a by-product of solving \eqref{equ:linear_system_weight}, no extra computation is needed to calculate the approximated $\Sigma_{\mathbf{\Delta}}'^{-1}$ using \eqref{equ:approximated_cov}. 

From the viewpoint of probability theory, the largest eigenvalue of $\Sigma_{\mathbf{\Delta}}$ reports the accuracy of the calculated estimates.

%\begin{flalign}
%\sum_{i=0}^{N} [ ( \mathbf{f}_i')^T \hat{ \mathbf{E}} \boldsymbol{\Sigma}_f \hat{ \mathbf{E}}^T \mathbf{f}_i' + \mathbf{f}_i^T \hat{ \mathbf{E}}^T \boldsymbol{\Sigma}_f \mathbf{E}\mathbf{f}_i ] = \sum_{i=0}^{N} \mathbf{J}_i \Sigma_{\Delta} \mathbf{J}_i^T,
%\end{flalign}
%where $\hat{ \mathbf{E}} = \lfloor \hat{\mathbf{t}} \times \rfloor \hat{\mathbf{R}}$ and $\Sigma_{\Delta}$ is the covariance of $\Delta$. The covariance of $(\delta \mathbf{f}_i')^T \lfloor \mathbf{t} \times \rfloor \mathbf{R} \delta \mathbf{f}_i$ is ignored since it is significantly small.

\subsection{Engineering Considerations}
%\begin{figure}[h]
%	\centering
%	\includegraphics[width=0.5\columnwidth]{grid}
%	\caption{A figure of grid that illustrates the feature matches collector. The grid width is $W$ and grid height is $H$ respectively. There are $W \times H$ cells in total that equally divide the whole image. Feature correspondences are collected in accordance with theirs left feature coordinates on the image.} 
%	\label{fig:grid}
%\end{figure}
Our proposed nonlinear optimization-based algorithm is backed by a carefully engineered system pipeline, as shown in Fig.~\ref{fig:framework}, to achieve high-precision calibration.
For each incoming image, we detect feature points on it using the ST corner detector~\cite{jianboCarlo94} and describe them using the BRIEF (binary robust independent elementary features) keypoint descriptor~\cite{beief}. 
We match features between the left and right images based on the distance metric in the BRIEF space. 
We collect a huge amount of features during the operation, and therefore we impose a strict outlier rejection scheme that consists of three steps:

\begin{itemize}
\item Prior information from the initial guess of the stereo extrinsic rejects outlier feature matches according to the epipolar error.
\item Matches from the left view to the right view and from the right view to the left view are checked for consistency, which is referred as a cross check. 
      The distance of the best match should also be less than one-third that of the second best match in the BRIEF space, and the matches are therefore subjected to a uniqueness check.
      Any features that fail either test are rejected.
\item Feature matches should share the same geometric model, such as the constraints of the fundamental matrix or essential matrix. Random sample consensus (RANSAC) \cite{ransac} is applied to further filter possible outliers.
\end{itemize}

We enforce even feature distribution by splitting the image into $W \times H$ equally-sized cells. The number of features within each cell is capped at $c_m$. This is necessary to ensure real-time operation.
Features with pixel locations within a cell are added or removed according to selection criteria as follows:

\begin{itemize}
	\item If the cell is not full, all matched features are kept and added;
	\item If the cell is full, $c_m$ feature matches with the largest disparities are kept and the others are dropped;
	\item If the cell is full and feature matches have similar disparities, $c_m$ features are kept randomly;
\end{itemize}

Note that this feature collection process is done only spatially in the image plane and does not require any temporal feature correspondences.
The reason that we save feature correspondences with large disparities is that these matches excite the error terms related to the translation between cameras, as well as the rotation perpendicular to the epipolar plane.
We experimentally verified that large-disparity features lead to better extrinsic calibration.
Feature correspondences that remain in the cells are used for the stereo extrinsic estimation described in Sect.~\ref{sec:opt}, followed by the covariance calculation in Sect.~\ref{sec:cov}. 
From the view of probability, we can identify that the accuracy of the extrinsic estimates terminates the calibration process if the largest eigenvalue of the covariance matrix is sufficiently small. 

%%%%%%%%%%%%%%%%%%%%%%%%%%%%%%%%%%%%%%%%%%%%%%%%%%%%%%%%%%%%%%%%%%%%%%%%%%%%%%%%%%%%%%%%%%%%%%%%%%%%%%%%%%%%%%%%%%%%%%%%%%5
\section{Experimental Results}
\label{sec:experiment}

\begin{figure}[h]
	\centering
	\includegraphics[width=0.5\columnwidth]{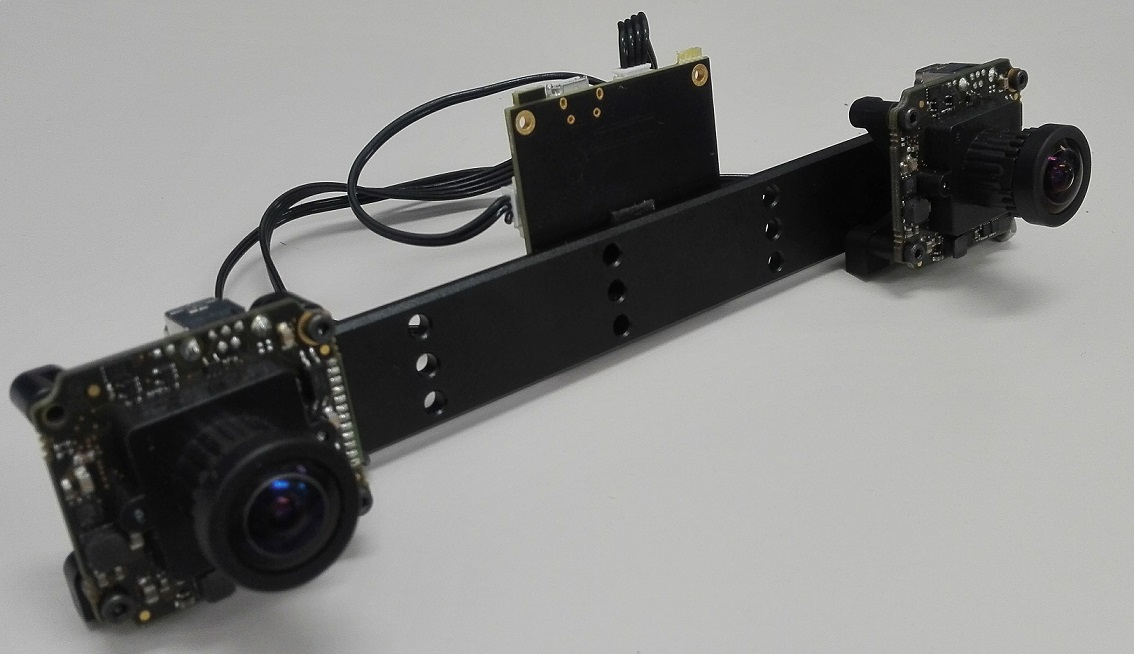}
	\caption{Our stereo setup with two hardware-synchronized cameras fixed on a metal bar. 
	         We pre-calibrate the cameras' intrinsic parameters.
	         The length of the baseline is assumed to be constant. 
	         We estimate the 5-DOF extrinsic parameters between the two cameras.} 
	\label{fig:stereo_baseline}
\end{figure}

\begin{figure*}[htp]
	\centering
	\subfloat[]{{\includegraphics[width=1.0\textwidth]{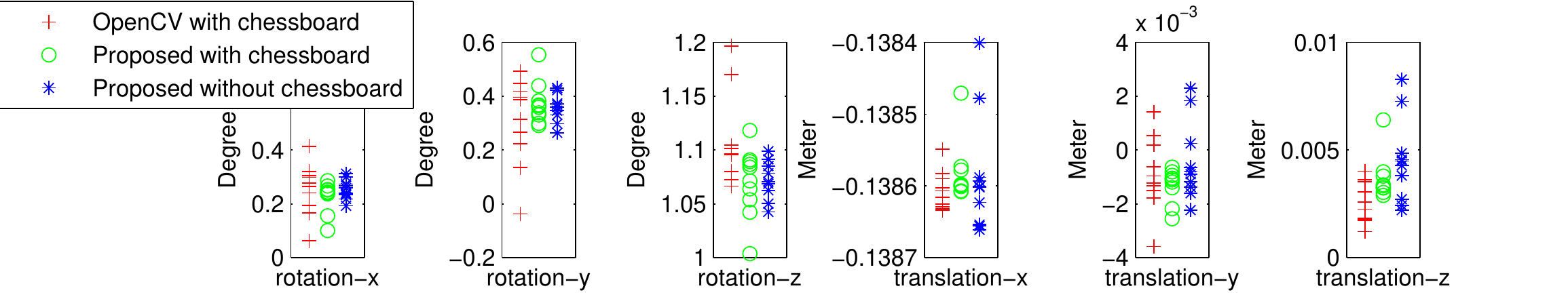}}} \\
	\caption{Statistics of estimated extrinsic obtained by OpenCV with a chessboard, our proposed method with a chessboard, and our proposed method without a chessboard. 
	         We perform ten trials for each of the scenarios. 
	         Images associated with different testing scenarios are shown in Fig.~\ref{fig:exp3_figs}.
	         We observe that in terms of both accuracy and repeatability, our markerless method provides results that are comparable to or even better than the standard chessboard-based method in OpenCV.}
	\label{fig:exp3}
\end{figure*}

\begin{figure*}[htp]
	\centering
	\subfloat[]{{\includegraphics[width=2.8cm]{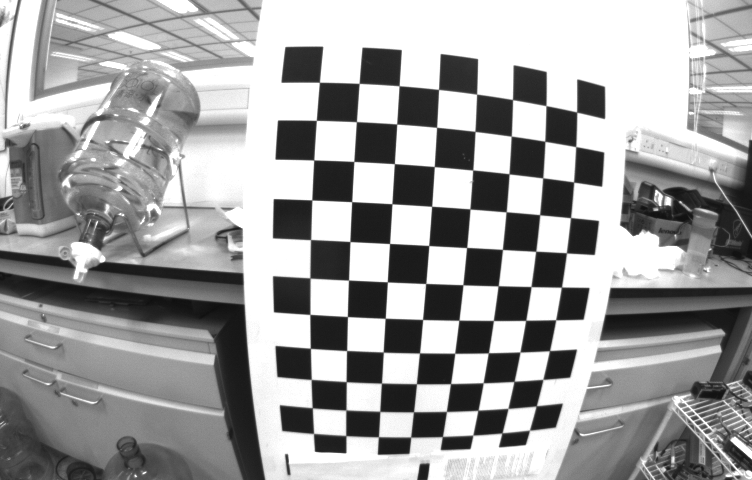} }}
	\subfloat[]{{\includegraphics[width=2.8cm]{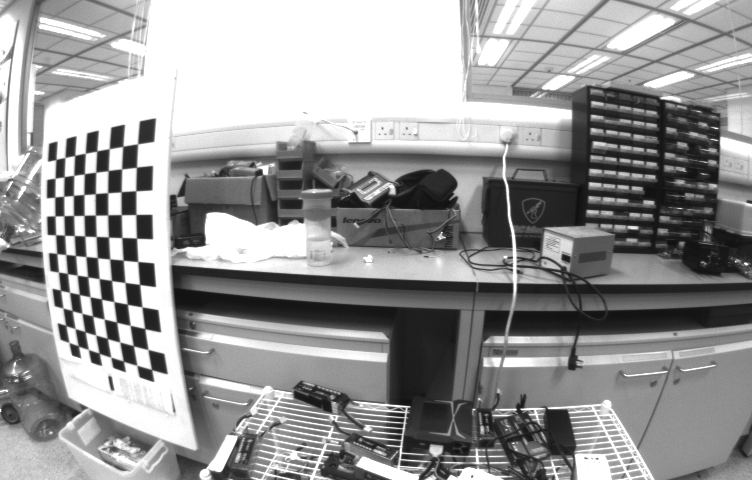} }} 
	\subfloat[]{{\includegraphics[width=2.8cm]{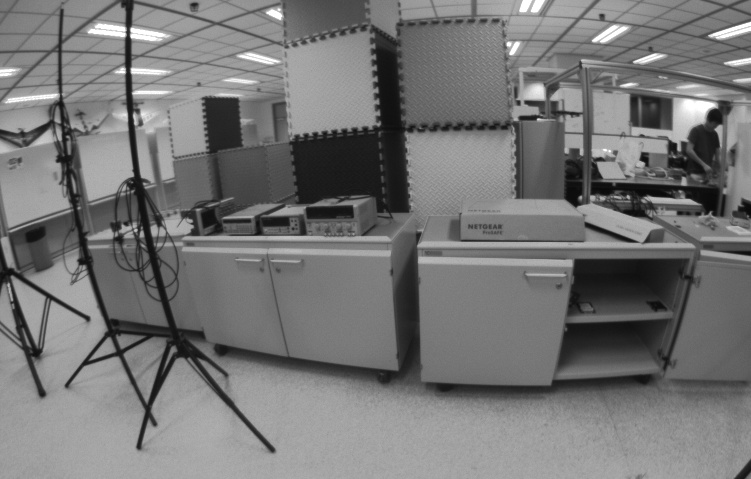} }}
	\subfloat[]{{\includegraphics[width=2.8cm]{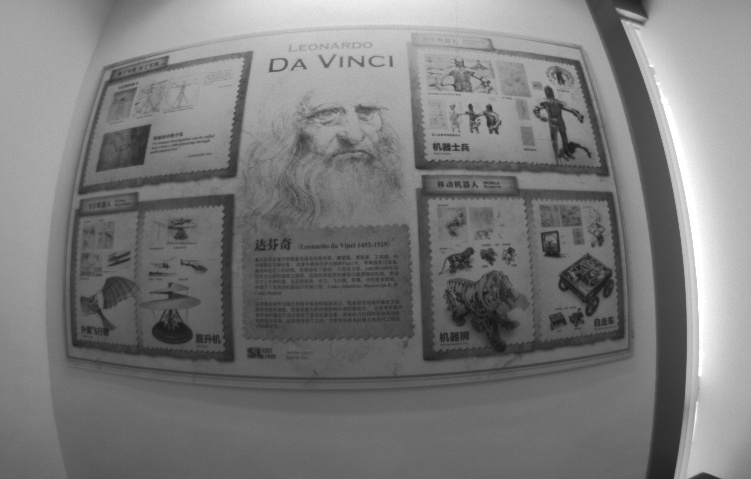} }} 
	\subfloat[]{{\includegraphics[width=2.8cm]{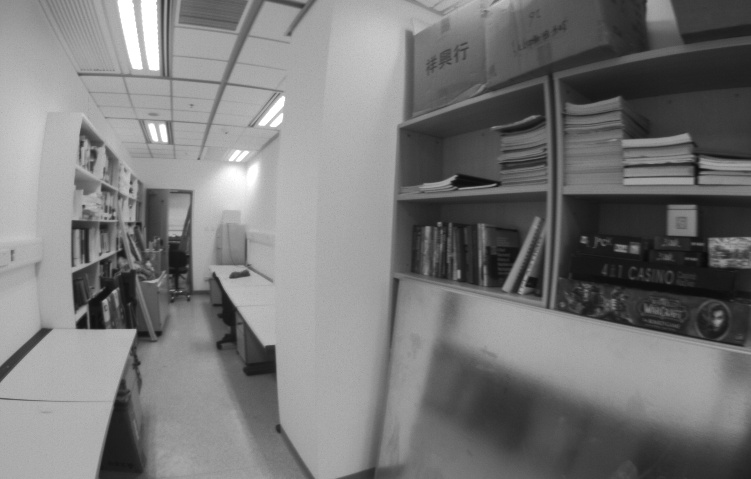} }}
	\subfloat[]{{\includegraphics[width=2.8cm]{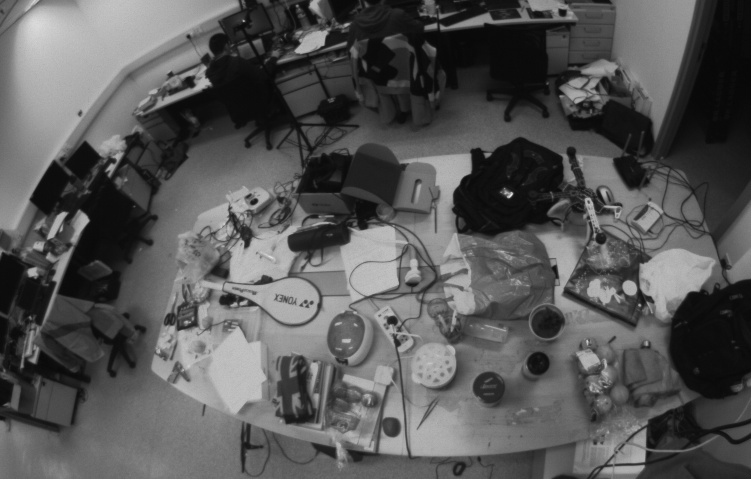} }} \\
	\subfloat[]{{\includegraphics[width=2.8cm]{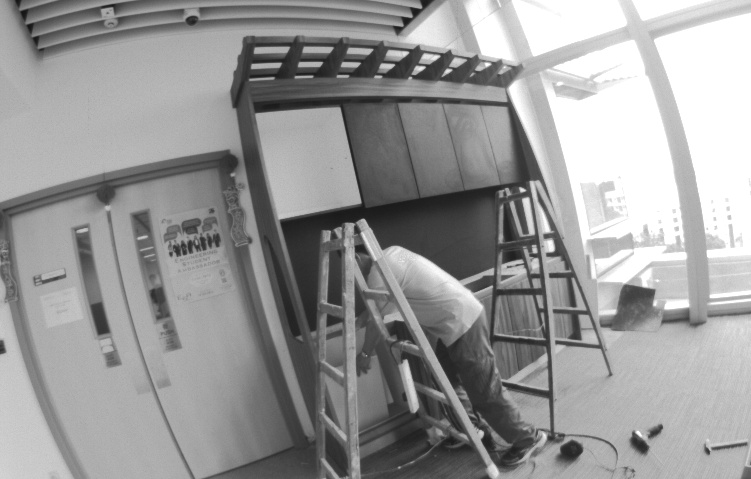} }}  
	\subfloat[]{{\includegraphics[width=2.8cm]{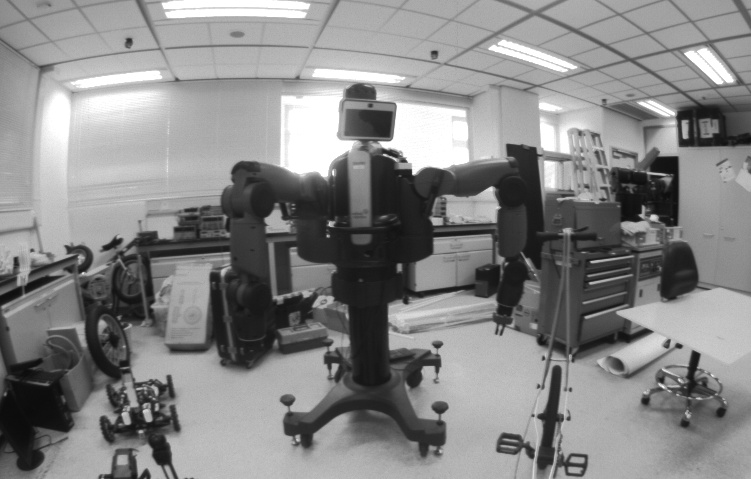} }}
	\subfloat[]{{\includegraphics[width=2.8cm]{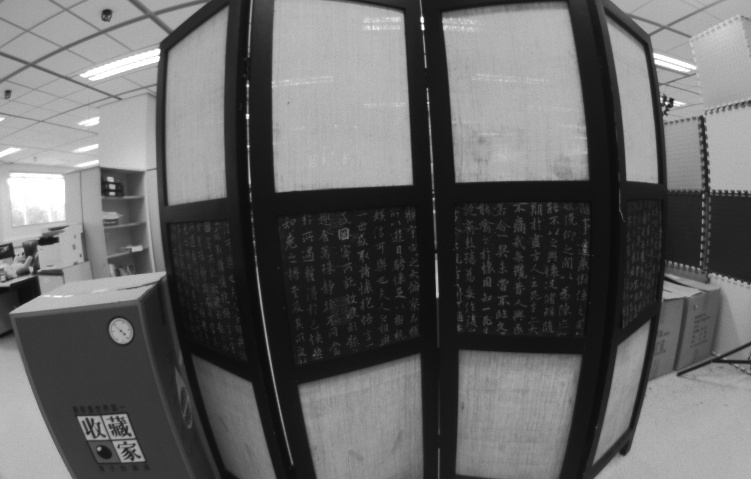} }} 
	\subfloat[]{{\includegraphics[width=2.8cm]{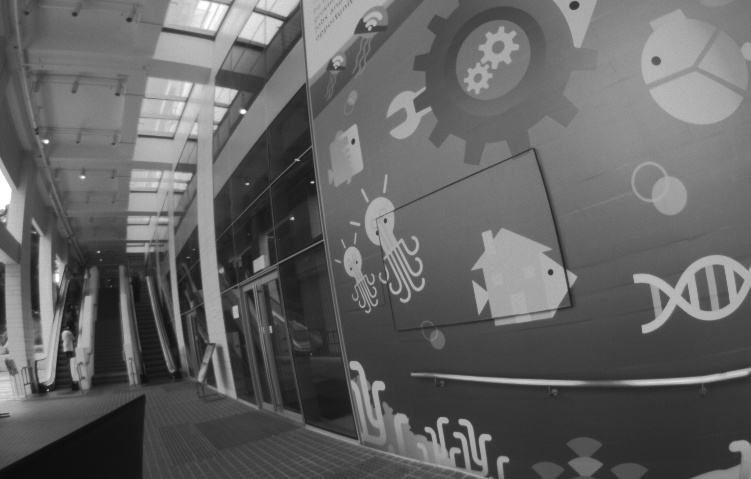} }}
	\subfloat[]{{\includegraphics[width=2.8cm]{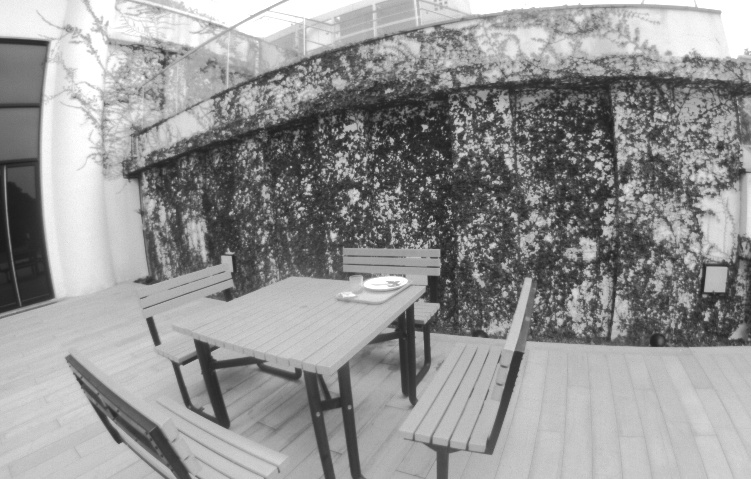} }}
	\subfloat[]{{\includegraphics[width=2.8cm]{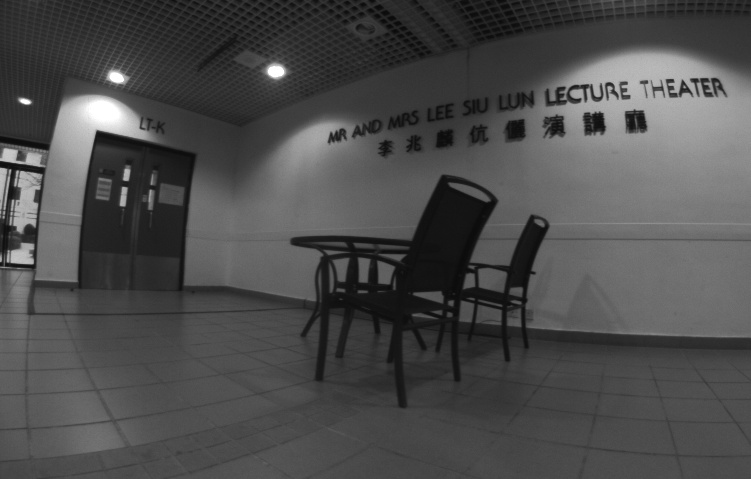} }}  \\
	\caption{(a) and (b) are sample images for marker-based calibration. 
	         (c)-(l) are representative images in the markerless case. We include indoor, outdoor, static, and dynamic scenes.
	         Calibration results are shown in Fig.~\ref{fig:exp3} and Table II.}
	\label{fig:exp3_figs}
\end{figure*}

\begin{table}[b]
	\label{tab:table_1}
	\centering
	\begin{tabular}{|c|c|}
		\hline
		Module & Average Computing Time  \\ 
		\hline
		Feature Detection & 20 ms  \\
		Feature Matching & 4 ms \\
		Feature Correspondence Filtering & 1 ms \\
	        Stereo Extrinsic Calibration & 10 ms \\
	        Covariance Calculation & 10 ms \\
		\hline
		Total & 45 ms\\
		\hline
	\end{tabular}
	\caption{Average computing time for each system module.}
\end{table}

\subsection{Implementation Details}
Our sensor suite consists of two global shutter cameras with a fronto-parallel stereo configuration, as shown in Fig.~\ref{fig:stereo_baseline}.  
All the experiments are run on a Lenovo Y510 laptop with i7-4720HQ CPU and $8\,\text{GB}$ of RAM. 
Our software is developed in C++ using OpenCV\footnote{\url{http://opencv.org/}}, Eigen\footnote{\url{http://eigen.tuxfamily.org/}} and ROS libraries.
The stereo cameras capture synchronized data at $10\,\text{Hz}$ at VGA resolution.
We use wide-angle cameras with a diagonal field of view of $140\,\text{degrees}$. The grid size in the feature matches collection module is $16 \times 25$ and the cell capacity is 10. 
Auto-exposure is enabled to account for illumination changes in indoor-outdoor scenes. 
The average computing time for each system module is highlighted in Table I. 

\begin{table}[b]
	\label{tab:table_2}
	\centering
	\begin{tabular}{|c|c|c|c|}
		\hline
		Mean/Std Dev&OpenCV & Ours-Marker &Ours-Markerless\\ 
		\hline
		rotation-x (deg)& 0.2541/0.0961 & 0.2271/0.0558& 0.2534/0.0366 \\
		rotation-y (deg)& 0.3041/0.1624 &  0.3722/0.0764 &  0.3621/0.0559\\
		rotation-z (deg)& 1.1084/0.0420 & 1.0704/0.0317 & 1.0739/0.0185 \\
		\hline
		translation-x (m)&  -0.1386/3e-05 & -0.1386/4e-05&  -0.1386/8e-05\\
		translation-y (m)& -0.0009/0.0014 &  -0.0013/6e-04&  -0.0004/0.0015\\
		translation-z (m)& 0.0026/0.0010 & 0.0037/0.0010 & 0.0045/0.0020\\
		\hline
	\end{tabular}
	\caption{The mean and standard deviation of extrinsic estimates from OpenCV and our approach both with and without chessboard. }
\end{table}

\begin{figure}[h]
	\centering
	\includegraphics[width=1.0\columnwidth]{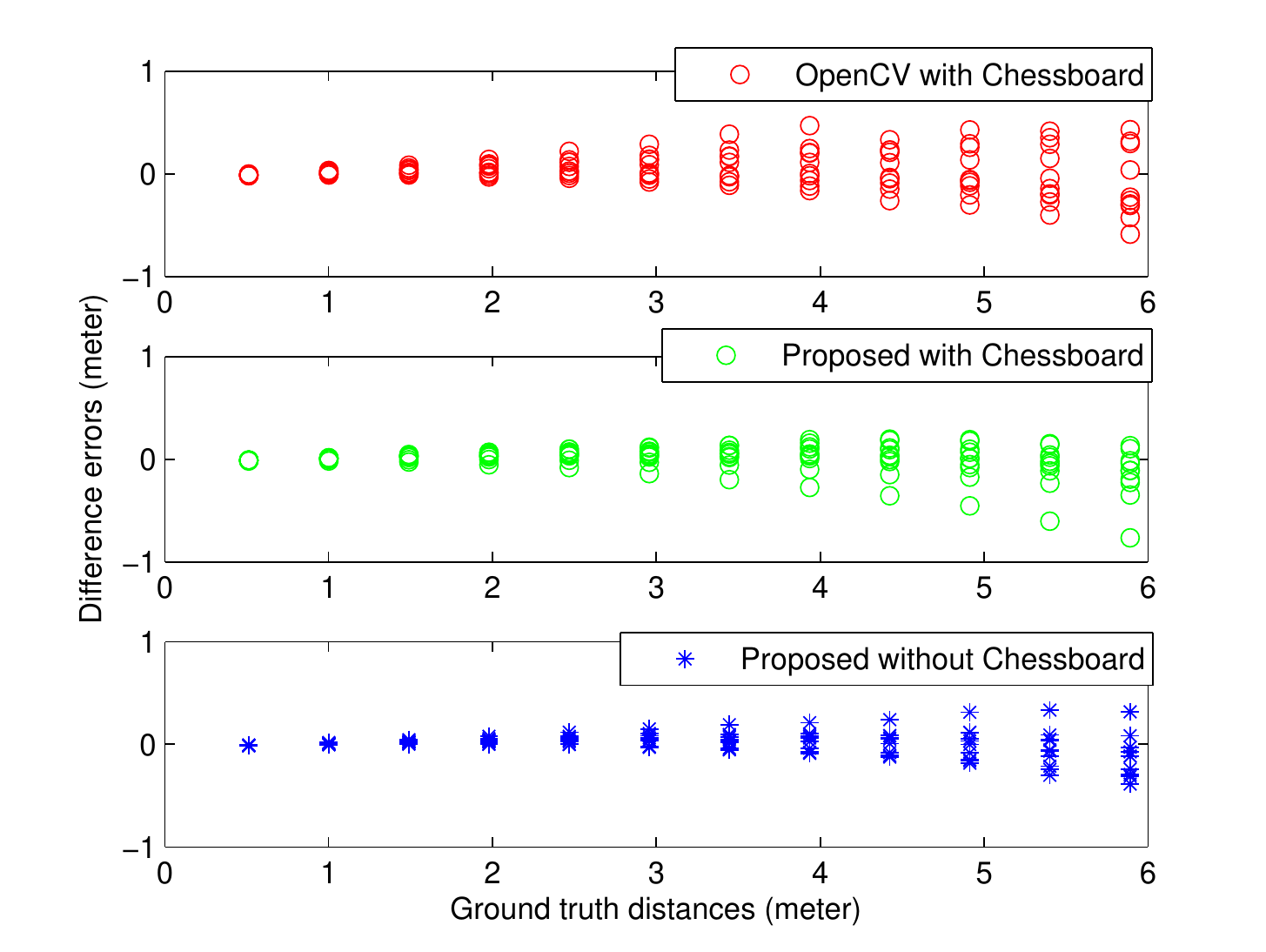} \\
	\caption{Distance measurement errors for the chessboard-based OpenCV toolbox, our proposed method with chessboard, and our proposed method with natural features.}
	\label{fig:dist}
\end{figure}

\begin{figure}[h]
	\centering
	\includegraphics[width=0.9\columnwidth]{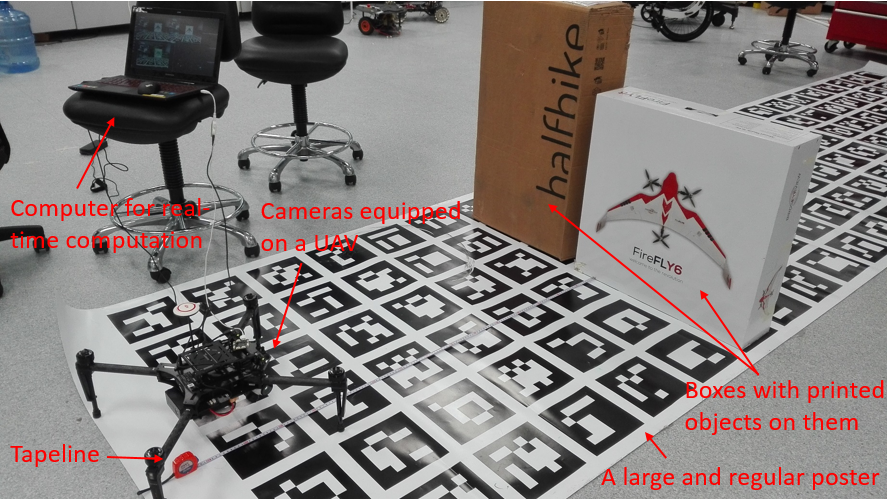} \\
	\caption{Experiment setup for the evaluation of distance measurements. }
	\label{fig:dist_setup}
\end{figure}

\begin{figure}[!ht]
	\centering
	\subfloat[]{{\includegraphics[width=2.8cm]{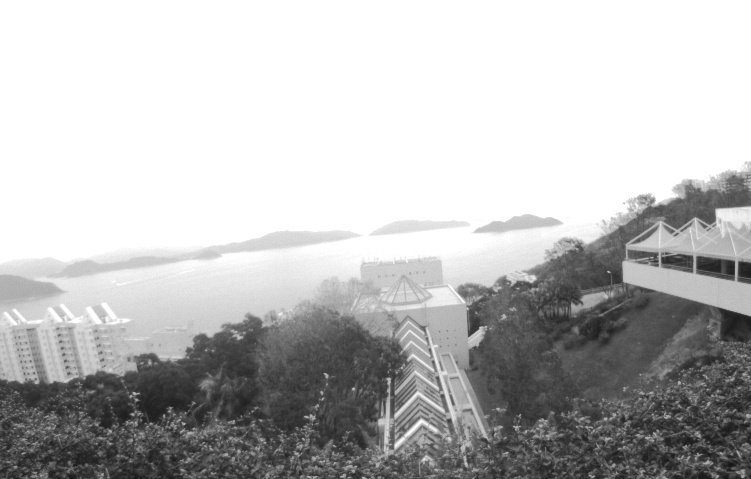} }}
	\subfloat[]{{\includegraphics[width=2.8cm]{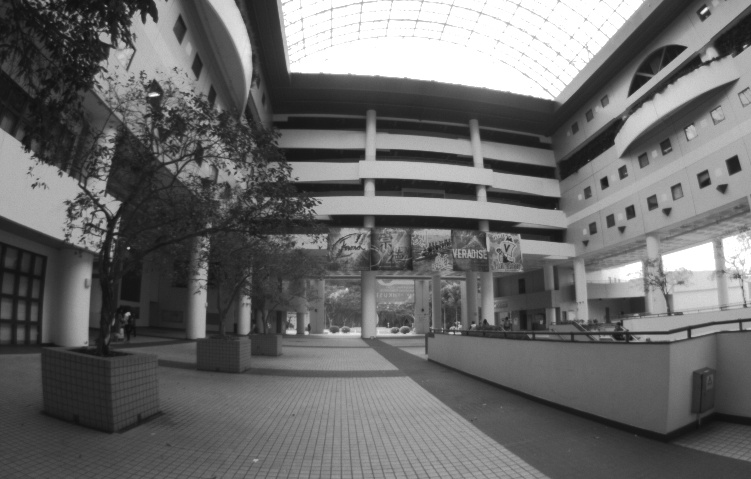} }} 
	\subfloat[]{{\includegraphics[width=2.8cm]{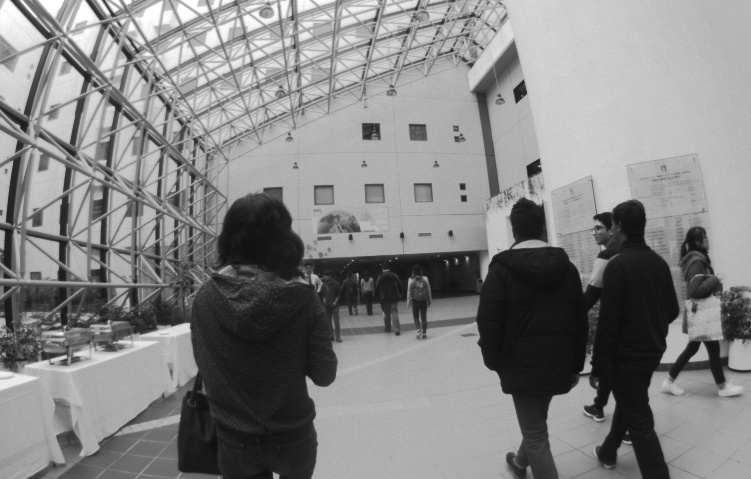} }} \\
	\subfloat[]{{\includegraphics[width=2.8cm]{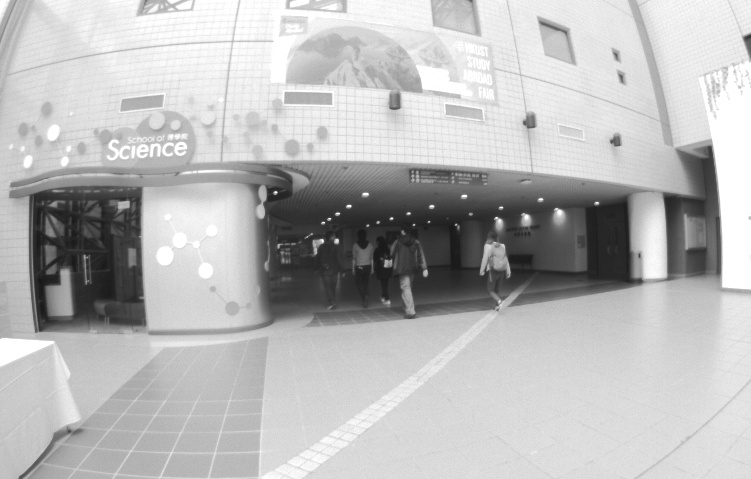} }} 
	\subfloat[]{{\includegraphics[width=2.8cm]{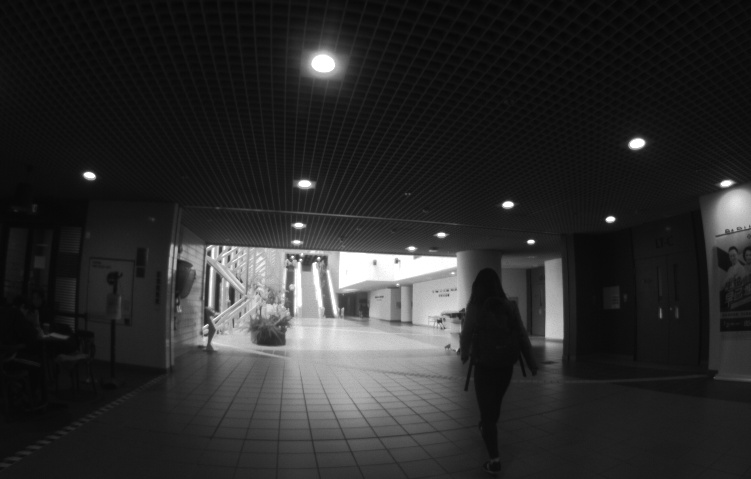} }}
	\subfloat[]{{\includegraphics[width=2.8cm]{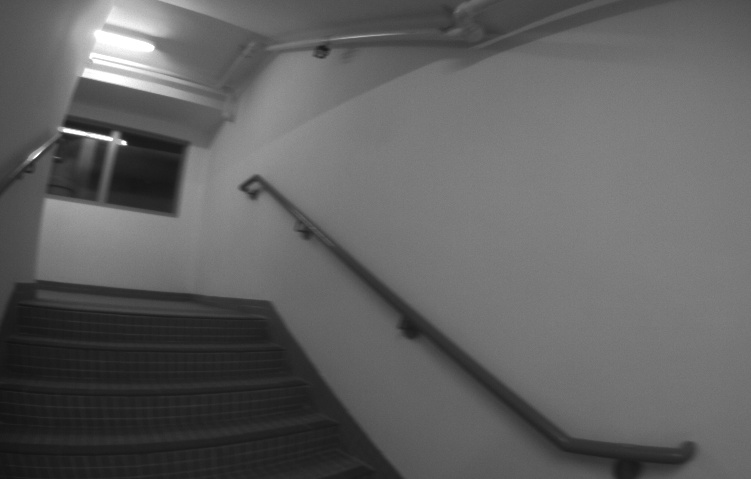} }} \\
	\subfloat[]{{\includegraphics[width=2.8cm]{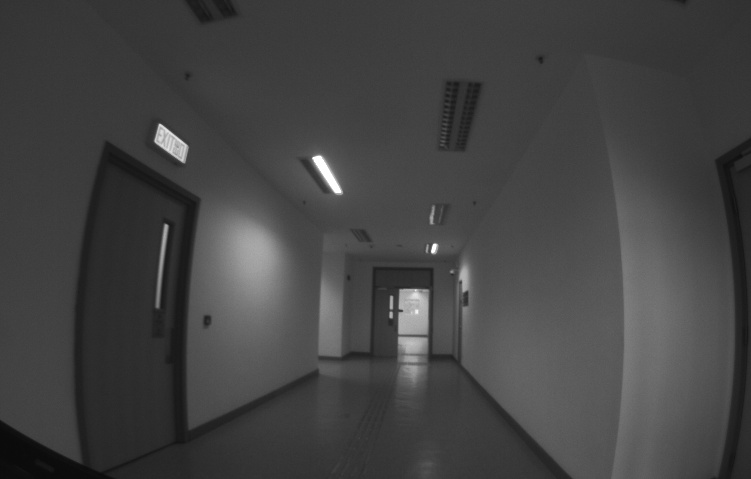} }}  
	\subfloat[]{{\includegraphics[width=2.8cm]{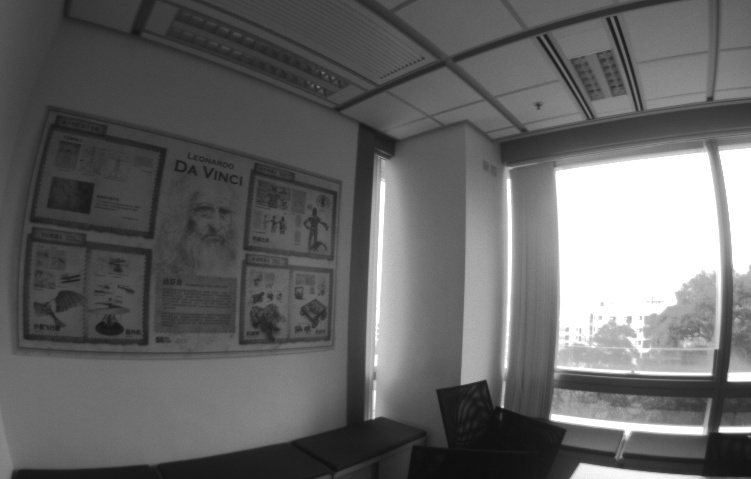} }}
	\caption{Snapshots during the continuous large-scale experiment. 
		Corresponding extrinsic estimates and standard deviations are shown in Fig.~\ref{fig:longterm}. }
	\label{fig:longterm_figs}
\end{figure}
\begin{figure}[!h]
	\centering
	\subfloat[Rotation]{{\includegraphics[width=0.9\columnwidth]{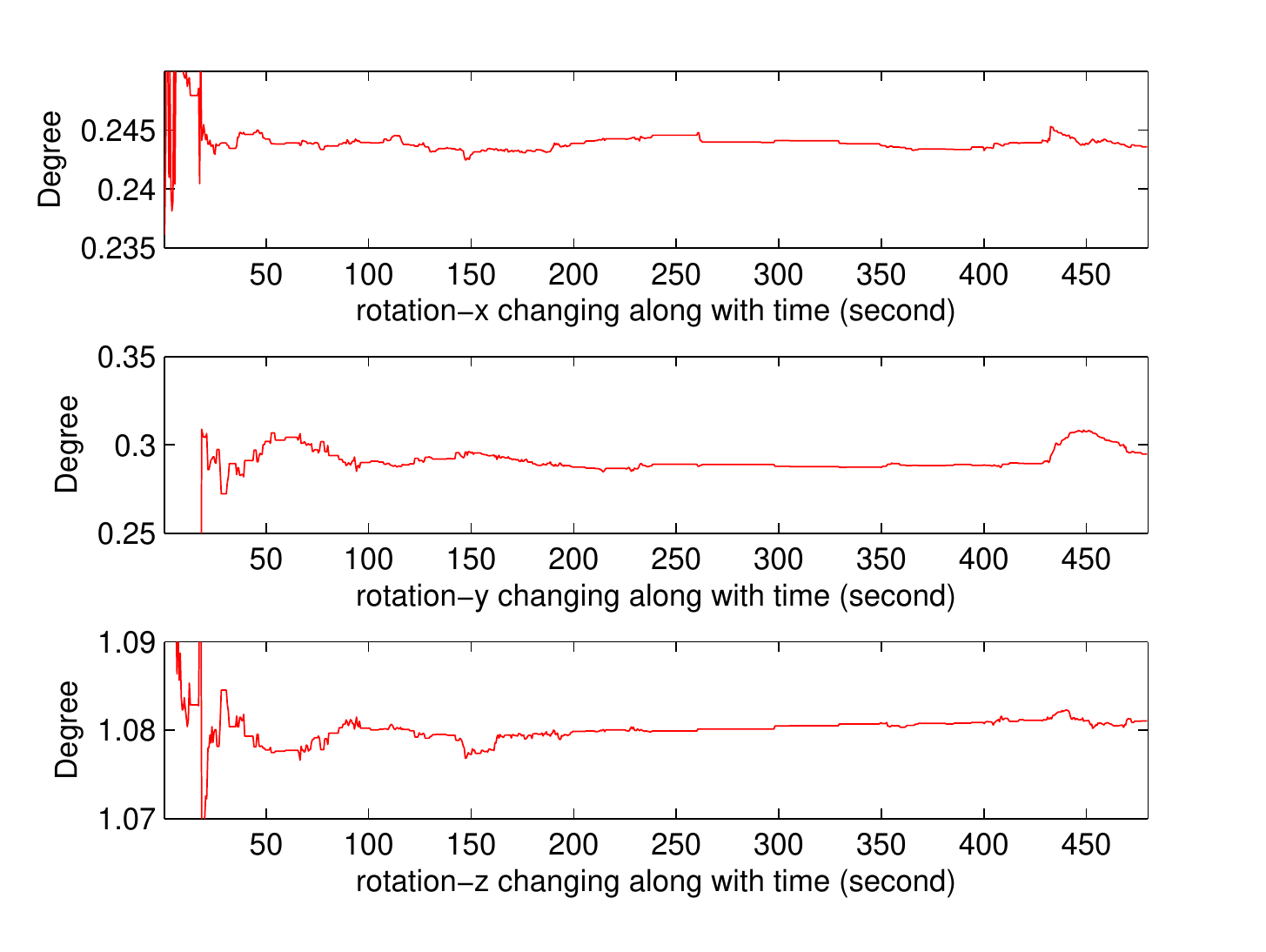}}}  \\
	\vspace{-10pt}
	\subfloat[Translation]{{\includegraphics[width=0.9\columnwidth]{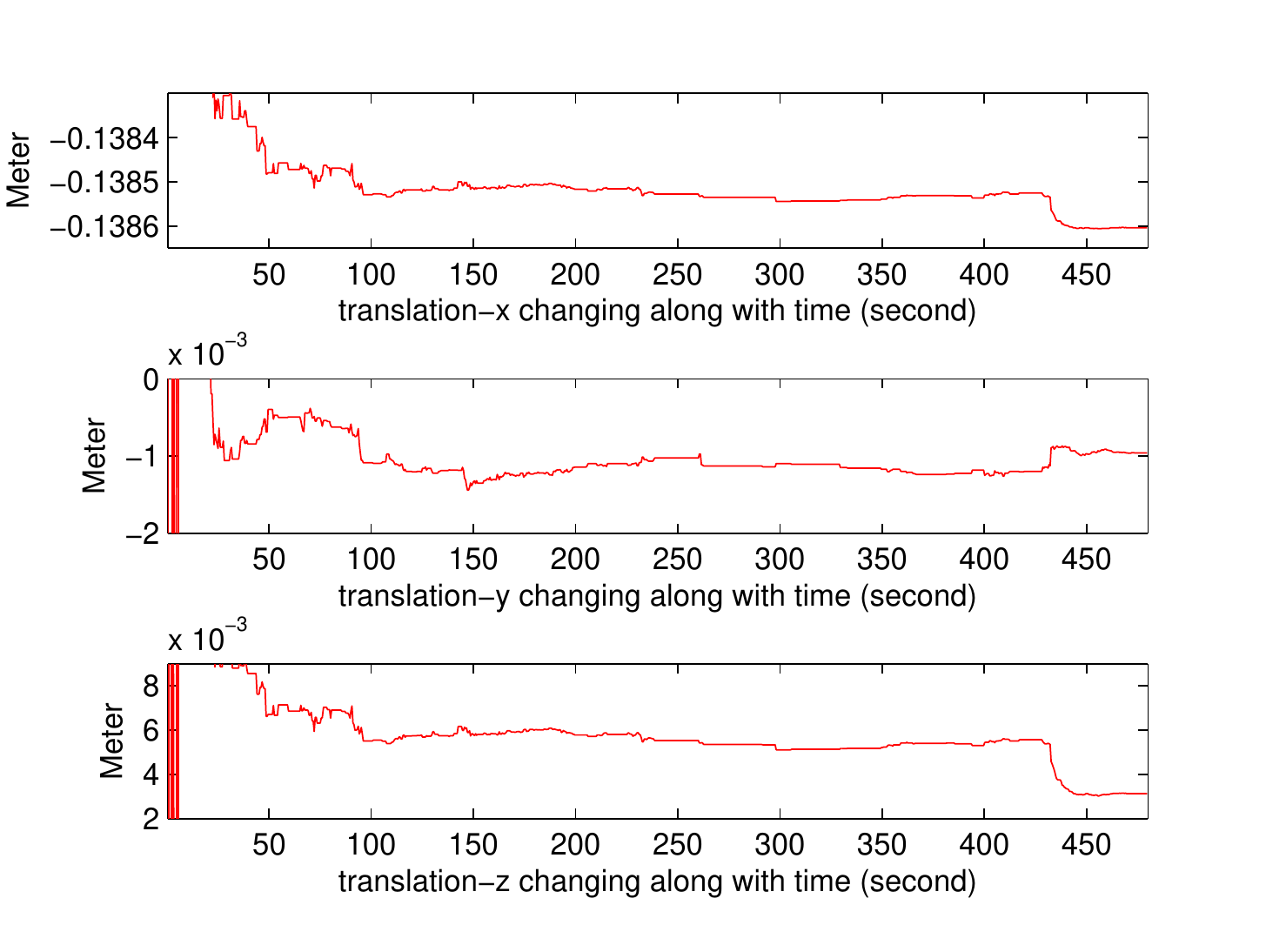}}} \\
	\vspace{-10pt}
	\subfloat[The log of the largest eigenvalue of the covariance matrix (Sect.~\ref{sec:cov}).]{{\includegraphics[width=0.9\columnwidth]{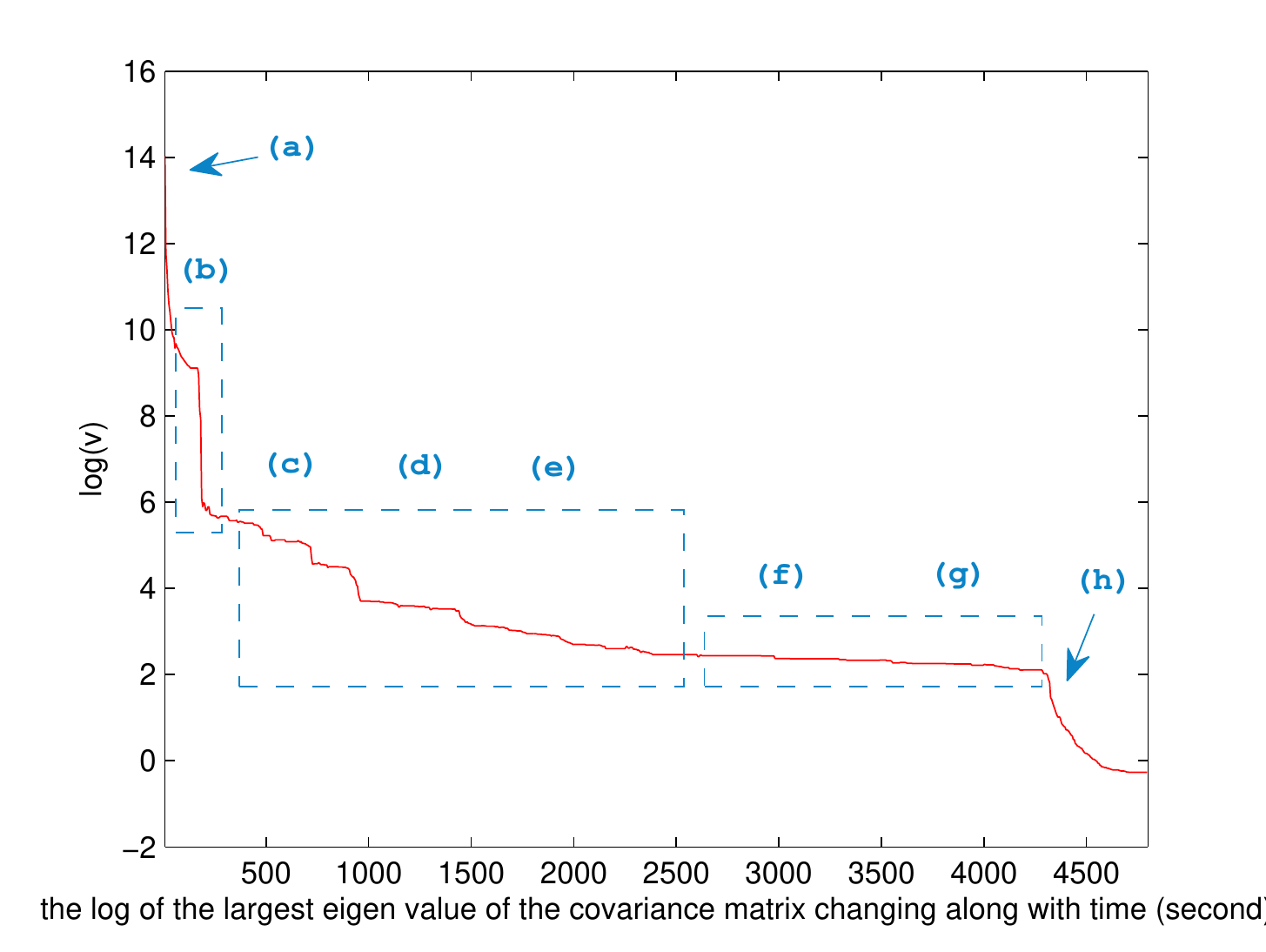}}} \\
	\caption{Statistics of a continuous stereo calibration experiment in a large-scale environment. 
		Convergence of all parameters is observed.
		We highlight that the largest eigenvalue of the covariance matrix changes according to the ``quality'' of the environment. 
		In distant scenes or textureless environments, the covariance does not change significantly.
		On the other hand, we see a fast covariance drop and fast parameter convergence in closed and texture-rich scenes.
		See the corresponding images in Fig.~\ref{fig:longterm_figs}. }
	\label{fig:longterm}
\end{figure}
\subsection{Comparison with Standard Offline Stereo Calibration in OpenCV}
In this experiment, we test whether our proposed approach can achieve an accuracy level that is comparable to the standard offline chessboard-based calibration toolbox in OpenCV. 
We conduct experiments in 3 settings: 1) OpenCV toolbox using a chessboard; 2) our method, but using a chessboard for feature detection and matching; and 3) our method with natural features.
Ten trials are conducted for each setting to verify the repeatability. 
For the first two cases, we follow the standard chessboard-based calibration process, where the board is observed from different angles, as shown in Fig.~\ref{fig:exp3_figs} (a) and (b). 
For the OpenCV method, we use the known size of the chessboard, while in our method we do not need this information because we assume the known length of the baseline.
For markerless calibration using our method, we perform tests in 10 different environments, including indoors, outdoors, and static and dynamic scenes, as shown in Fig.~\ref{fig:exp3_figs} (c)-(l). 
For fair comparison, the cameras' intrinsic parameters are pre-calibrated through marker-based monocular camera calibration and fixed for all the experiments. 
We convert the relative rotations $\mathbf{R}$ to Euler angels for graphical comparison, with the statistics shown in Fig.~\ref{fig:exp3} and Table II.
Results from the OpenCV toolbox are not great, due to the use of wide angle cameras.
We do observe larger deviations in the rotation around the cameras' y-axis, which is the rotation perpendicular to the epipolar plane. 
This makes sense as rotation along this axis is harder to observe from the epipolar constraints.

The ultimate goal of stereo extrinsic calibration is to improve dense stereo mapping and offer accurate distance measurements. 
We conduct another experiment to compare the distances calculated using the estimates from the OpenCV method, our proposed method with a chessboard, and our proposed method without a chessboard. 
The experiment setup is shown in Fig.~\ref{fig:dist_setup}, with the stereo cameras installed on a UAV.
We compute the distance between the cameras and a texture-rich printed object using the standard block matching dense stereo algorithm in OpenCV.
We conduct the test in multiple hand-measured ground-truth distances, and conduct distance measurements using all extrinsic parameters shown in Fig.~\ref{fig:exp3} for all 3 calibration settings.
Results are presented in Fig.~\ref{fig:dist}, and they suggest that our method achieves distance measurement accuracy that is comparable to or even better than that of the marker-based OpenCV toolbox.

Finally, in Fig.~\ref{fig:calibration}, we show qualitative results of dense stereo matching using extrinsic parameters from our markerless calibration method.
We see that our method significantly improves the stereo matching quality compared to the initially incorrect stereo extrinsic parameters.

\subsection{Long-term Real-time Stereo Calibration in Large-Scale Environments}
We demonstrate the performance of our markerless calibration system in a 9 minute continuous calibration experiment in a large-scale environments.
The estimated angles, translations and the logarithm of the largest eigenvalue of the covariance matrix (Sect.~\ref{sec:cov}) are shown in Fig.~\ref{fig:longterm}.
Representative images are shown in Fig.~\ref{fig:longterm_figs}. 

In this experiment, we first face a scene with only far-away objects (Fig.~\ref{fig:longterm_figs} (a)). 
The standard deviation remains large at this moment. 
We turn around and walk into a campus building, and more close-by features are observed (Fig.~\ref{fig:longterm_figs} (b)). 
The standard deviation therefore decreases rapidly. 
In the following stage, we encounter pedestrians, changing lighting conditions, and transitions between indoor and outdoor environments (Fig.~\ref{fig:longterm_figs} (c)-(e)). 
Our algorithm works under all these situations and aggregates more and more feature matches for the stereo extrinsic calibration. 
Standard deviation decreases accordingly. 
We then go through featureless stairs and corridors, where the estimated covariance remains almost unchanged (Fig.~\ref{fig:longterm_figs} (f), (g)). 
However, as we step into a room facing a poster that is full of salient features, the covariance decreases dramatically (Fig.~\ref{fig:longterm_figs} (h)). 
Changes in the logarithm of the largest eigenvalue of the covariance matrix with respect to different situations are illustrated in Fig.~\ref{fig:longterm} (c).

\section{Conclusion and Future Work}
\label{sec:conclusions}
In this work, we propose a novel markerless stereo extrinsic estimation method based on 5-DOF nonlinear optimization on a manifold.
Our method captures the true sources of error for stereo extrinsic. It also comes with a principled covariance estimation method to identify estimator convergence.
We combine the theoretical results with careful engineering decisions on the system pipeline and obtain results that are comparable to standard offline chessboard-based stereo calibration methods. 
Our method enables reliable dense stereo matching for long-term autonomy.
In the next stage, we will extend this work to refine the cameras' intrinsic and extrinsic parameters simultaneously.

\bibliographystyle{IEEEtran}

\end{document}